\documentclass{article}

\usepackage[final]{corl_2019} % Uncomment for the camera-ready ``final'' version
\usepackage{amsmath,amsfonts,amsthm,bm}
\usepackage{graphicx}
\usepackage{listings}
\usepackage{courier}
\usepackage{wrapfig}
\usepackage{enumitem}
\usepackage{multicol}

\title{Multi-Agent Manipulation via Locomotion \\using Hierarchical Sim2Real}

\author{
  Ofir Nachum\\
  Google AI\\
  \texttt{ofirnachum@google.com} \\
  \And
  Michael Ahn\\
  Google AI\\
  \texttt{michaelahn@google.com} \\
  \And
  Hugo Ponte\thanks{Work done as a member of the Google AI Residency program (\url{g.co/airesidency}).}\\
  Google AI\\
  \texttt{hugo.p.cmu@gmail.com} \\
  \And
  Shixiang (Shane) Gu\\
  Google AI\\
  \texttt{shanegu@google.com} \\
  \And
  Vikash Kumar\\
  Google AI\\
  \texttt{vikashplus@gmail.com} \\
}

\def\pihi{\pi_{\mathrm{hi}}}
\def\pilo{\pi_{\mathrm{lo}}}
\def\rlo{r_{\mathrm{lo}}}
\def\rhi{r_{\mathrm{hi}}}
\def\raux{r_{\mathrm{aux}}}
\def\ahi{a_{\mathrm{hi}}}

\def\G{\mathcal{G}}
\def\uniform{\mathrm{uniform}}

\begin{document}
\maketitle

\setcounter{footnote}{0}

%===============================================================================

\begin{abstract}
Manipulation and locomotion are closely related problems that are often studied in isolation. In this work, we study the problem of coordinating multiple mobile agents to exhibit manipulation behaviors using a reinforcement learning (RL) approach. 
Our method hinges on the use of {\em hierarchical sim2real} -- a simulated environment is used to learn low-level goal-reaching skills, which are then used as the action space for a high-level RL controller, also trained in simulation. 
The full hierarchical policy is then transferred to the real world in a zero-shot fashion. 
The application of domain randomization during training enables the learned behaviors to generalize to real-world settings, 
while the use of hierarchy provides a modular paradigm for learning and transferring increasingly complex behaviors.
We evaluate our method on a number of real-world tasks, including coordinated object manipulation in a multi-agent setting.\footnote{See videos at \url{https://sites.google.com/view/manipulation-via-locomotion}.}

\end{abstract}

 
\section{Introduction}

%VK-6-28: Complex behaviors resulting from co-ordianted movements between various group members have enabled various species to achieve far great feat than individual members were other wise capable -- a colony of ants can carry foods multiple time their own body weight. This is one of the primary reason behid species evolving to live in groups. The complexities of the behavior exhibited by co-ordinating agents are far too complex to be classified as either primarily manipulation or primarily locomotion. For example consider a group of lions attacking a prey elephant. Are they navigating with elephant being in the way? Or are they manipulating an elephant for a kill? When we walk -- are we navigating forward on earth or are we manipulating the earth under us via our feets?

%Manipulation and locotion, which are often studied in isolation, are indeed dual to each other[]. In this work we study the duality of the problem. We propose a hierirchial view point to the dual problem and propose ... 

%We have seen progress in locomotion paragraph 

%We have seen progress in manipulation paragraph

%Sim2Real has made significant stride individually in both Manipulation and Locomotion. However when you study the fused dual problem of manipulation and locomotion  -- to much to randomize (and follow along with your box xample )

The coordinated effort of a group allows for far greater results than any individual could ever hope to achieve; e.g., a colony of ants can work together to build intricate underground structures. This is one of the primary reasons behind species evolving to live in groups~\cite{alexander1974evolution}. 
The behaviors exhibited by coordinating agents are often far too complex to be classified as either primarily manipulation or primarily locomotion. For example, consider a group of lions attacking a prey zebra. Are they navigating to the vulnerable spots of the zebra? Or are they manipulating the zebra for a kill?  When we walk -- are we navigating forward in the world, or are we manipulating the Earth via our feet? %earth under us via our feets?

\begin{wrapfigure}{R}{0.32\textwidth}
\begin{minipage}{0.32\textwidth}
\centering
\vspace{-0.15in}
\includegraphics[width=0.99\columnwidth,height=0.92\columnwidth]{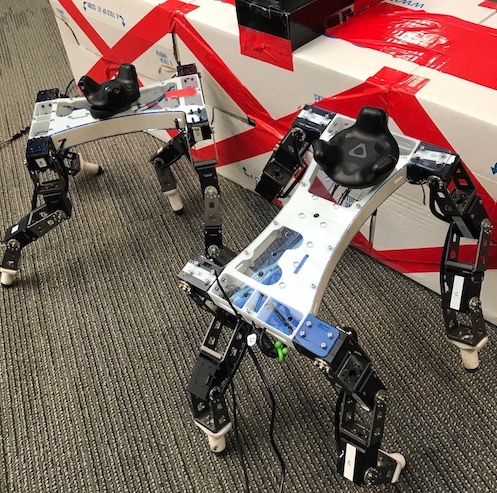}
\caption{Multiple agents coordinate to move a large object.}
\vspace{-0.25in}
\label{fig:intro}
\end{minipage}
\end{wrapfigure}

In this work, we aim to enable legged robots to exhibit these {\em manipulation via locomotion} behaviors 
by use of model-free reinforcement learning (RL).
In contrast to traditional approaches relying on manual hand-design or detailed understanding of system dynamics, the use of model-free RL has the potential to provide more robust policies, as it requires little domain knowledge other than access to a reward signal. This makes it especially desirable in domains where dynamics are difficult to model, such as diverse environments (e.g., an indoor office terrain vs. an outdoor natural terrain)~\cite{learn2walk} and in the presence of complex or faulty hardware.
While recent works have made great progress in applying RL to learn robust locomotion policies on real-world legged robots~\cite{kohl2004policy,haarnoja2018soft,tan2018sim}, 
examples of using RL on legged robots for coordination or manipulation remain elusive.
%In contrast to traditional approaches relying on manual hand-design or detailed understanding of system dynamics, the use of model-free RL has the potential to provide more robust policies, as it requires little domain knowledge other than access to a reward signal. This makes it especially desirable in domains whose dynamics are difficult to model, such as diverse environments (e.g., an indoor office terrain vs. an outdoor natural terrain)~\cite{learn2walk} and in the presence of complex or faulty hardware.
%
%Despite these recent real-world successes, examples of RL for learning behaviors more complex than simple walking or path following remain elusive.  
%Despite these recent real-world success, examples of legged robots using RL for learning how to {\em manipulate} their environments rather than for learning simple {\em locomotion} (walking or path following) remain elusive.

%In part, this is due to the reliance of modern RL algorithms on large amounts of trial-and-error experience, with larger amounts necessary for more complex tasks.  In the real-world, this requirement is prohibitively expensive to the point of being infeasible for most tasks.
In part, this is due to the increased complexity of agent-object and agent-agent interactions, which exacerbates the reliance of modern RL algorithms on large amounts of real-world trial-and-error experience.
%Recent works have proposed to bypass this bottleneck by training policies in simulation and accounting for unknown differences between real-world and simulated dynamics via domain randomization~\cite{sadeghi2016cad2rl,tobin2017domain,andrychowicz2018learning,matas2018sim}.
%However, domain randomization has its own scalability challenges.
While recent works have proposed to bypass the experience bottleneck by training policies in simulation and accounting for unknown differences between real-world and simulated dynamics ({\em sim2real}) via domain randomization~\cite{sadeghi2016cad2rl,tobin2017domain,andrychowicz2018learning,matas2018sim},
domain randomization has its own scalability challenges.
As a robotic task becomes more complex, its simulated counterpart must utilize more randomizations; e.g., a walking task may only need to randomize properties associated with the robot joint dynamics, but if objects are introduced to the scene, the object properties must be randomized as well.
%Methods for choosing and tuning these randomizations are manual or otherwise expensive to perform~\cite{mehta2019active}.
%Moreover, if 
In addition to the scalibility issues associated with appropriately tuning these randomizations~\cite{mehta2019active}, if too many randomizations are used, the optimal policy found in simulation may be too conservative~\cite{ramos2019bayessim}, to the point of being unable to adequately solve either the simulated or the original real-world task.

We propose to tackle these scalability challenges and enable legged robots (specifically in this paper, quadrupeds) to learn coordination and manipulation behaviors by introducing hierarchy.
%We propose to tackle the real-world scalability challenges associated with model-free RL by introducing hierarchy.
Simulation results have shown that introducing hierarchy can elevate the use of RL from simple locomotion to complex navigation and object manipulation~\cite{hiro}.
Thus, we propose doing the same for real-world settings. We separate policies into low-level {\em goal-conditioned} policies that are trained to perform locomotive behavior, and high-level {\em goal-proposing} policies that are trained to solve a task by iteratively 
directing the locomotion of the low-level policy.
To properly scale the use of domain randomization, %and in contrast to similar work~\cite{jain2019hierarchical,hiro}, 
we propose to train each level of the hierarchy successively rather than concurrently.  That is, the low-level policy is first trained with domain randomization chosen to be appropriate for its own goal-reaching task; the low-level is then fixed and the high-level policy is trained with randomization chosen to be appropriate for its task.
The modularity introduced by hierarchy is thus key to scaling up domain randomization, since each level of the hierarchy need only be trained with randomizations appropriate for its level of abstraction, and these are often much fewer (and thus less aggressive) than the randomizations necessary when training a single policy from scratch.
We call our training paradigm {\em hierarchical sim2real}.

\begin{figure}
  \setlength{\tabcolsep}{0pt}
  \renewcommand{\arraystretch}{0.8}
  \begin{center}
  \vspace{-0.1in}
  	\begin{tabular}{c}
  	\hspace{0.02\textwidth} Avoid \hspace{0.24\textwidth} Push \hspace{0.24\textwidth} Coordinate\\
  	\includegraphics[width=0.99\textwidth]{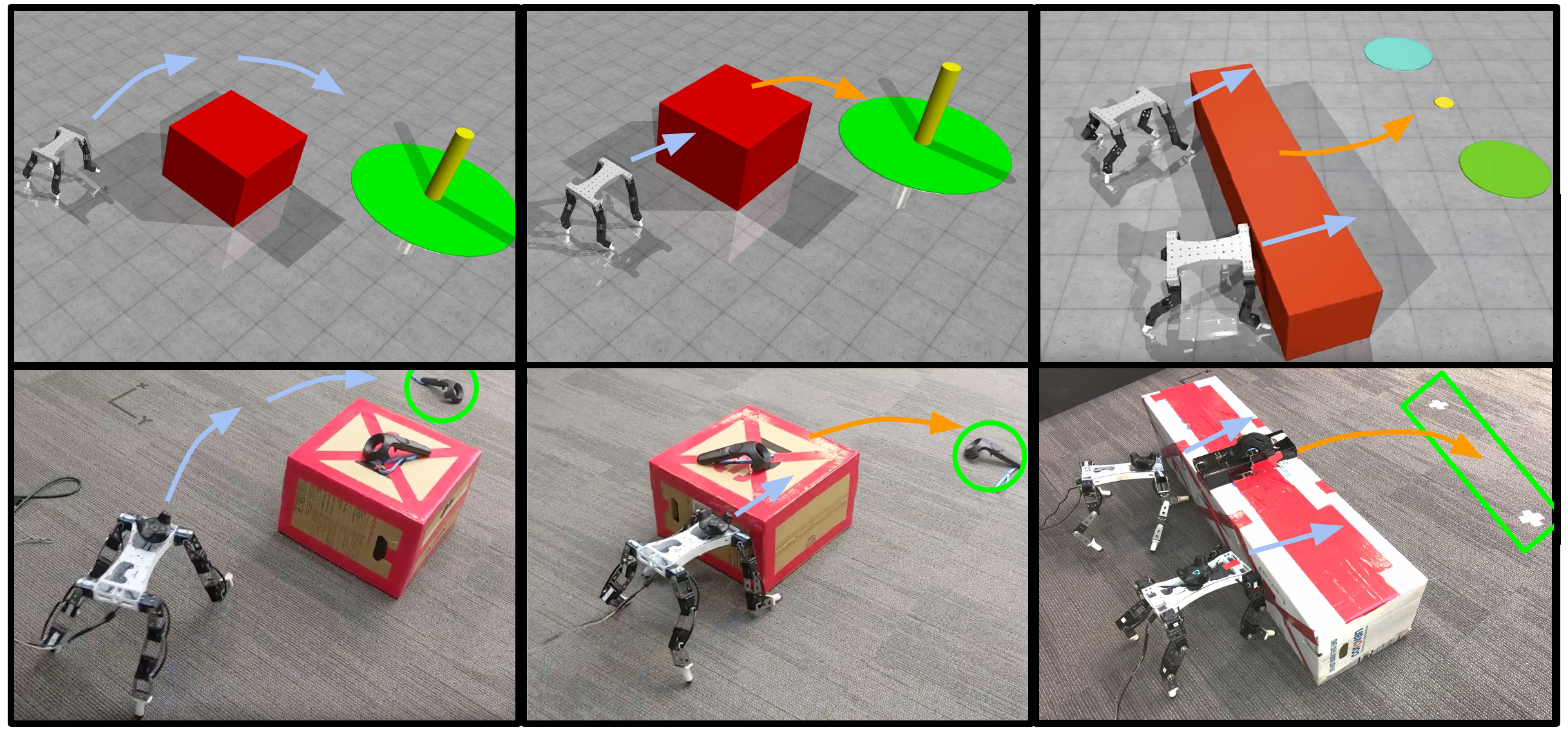}
  	\end{tabular}
  \end{center}
  \vspace{-0.1in}
  \caption{We consider three quadrupedal locomotion tasks of increasing complexity, utilizing the \textbf{D'Kitty} robot (see Section~\ref{sec:hardware} for details on this robot).  From left to right, we present the simulated (top row, using MuJoCo~\cite{mujoco}) and real-world (bottom row) versions of the three tasks: \textit{Avoid}, in which the quadruped must walk to a target location while avoiding a block object; \textit{Push}, in which a quadruped must push a block object to a desired location; and \textit{Coordinate}, in which two quadrupeds coordinate to push a long block to a target location and orientation. We utilize HTC Vive controllers and trackers to track the real-world position and orientation of agents, objects, and (for \textit{Avoid} and \textit{Push}) the desired target locations.}
  \vspace{-0.3in}
	\label{fig:tasks}
\end{figure}

Hierarchical sim2real lends itself well to legged robot tasks, in which it is natural to first learn a low-level policy to perform basic locomotion.  Given a well-performing locomotive policy, a high-level policy can then perform much more complex tasks, effectively treating the robot as a simple point mass.
Accordingly, we apply hierarchical sim2real to a number of challenging real-world quadrupedal tasks,
going beyond simple locomotion to demonstrate zero-shot transfer of learned policies for object avoidance and targeted object pushing. 
Ablation studies provide insight into our method and show the necessity of using both hierarchy and domain randomization in order to achieve good and efficient performance.
Finally, our results culminate in the successful demonstration of a challenging multi-agent task, in which two quadrupeds must coordinate their movement to push a heavy block to a desired target location and orientation.
This way, we are able to show the use of locomotion to perform coordinated multi-agent manipulation.

\section{Related Work}
\label{sec:related}
Before elaborating on the specifics of our method, we review previous works looking at manipulation via locomotion, hierarchy, and sim2real.
%Our method combines hierarchy and sim2real via domain randomization to demonstrate zero-shot transfer of complex reinforcement learned behaviors to real-world quadrupedal robots.  We review previous work in these fields.

\paragraph{Multi-Agent Manipulation via Locomotion}
The dual relationship between manipulation and locomotion~\cite{asfour2018dualities,mordatch2012contact} is evident in several previous works using legged robots for object manipulation~\cite{silva2018crabot}.
Much work has continued to develop such behaviors in the presence of multiple agents; see~\cite{tuci2018cooperative} for a comprehensive review on such {\em multi-robot systems}. For example, \cite{mataric1995cooperative} assess the ability of two robots to coordinate their behavior to transport an elongated box, similar to the \textit{Coordinate} task we consider. 
While several later works aimed at similar coordination problems use RL to some extent~\cite{olivares2004line,sanz2008applying,wang2006multi}, we believe our work is the first to use RL for the manipulation via locomotion problem as a whole. Our policy operates directly on the joint motors of the robot and is imbued with minimal domain knowledge, whereas previous works often restrict the use of RL to only high-level planning.

\paragraph{Hierarchy}
Hierarchy is key to the success of our method.  Our hierarchical decomposition of quadrupedal tasks is natural, and is inspired by similar decompositions in the robotics literature.  
Many previous successful examples of quadrupedal or humanoid locomotion rely on a hierarchical decomposition of the problem, in which a high-level controller optimizes for path or task space trajectories while a low-level controller optimizes joint movement and foot placement~\cite{feng2016online,zucker2010optimization,hutter2014quadrupedal,sentis2005synthesis,kalakrishnan2010fast,herzog2016momentum}.
These previous works optimize each controller in the hierarchy using methods such as inverse kinematics~\cite{feng2016online}, inverse dynamics~\cite{herzog2014balancing}, or planning~\cite{zucker2010optimization}.
In contrast, the focus of our method is on the use of model-free RL to optimize policies, which is potentially more scalable as it requires relatively little knowledge of the environment dynamics.
While hierarchical model-free RL has enjoyed much recent success in simulation~\cite{hiro,nachum2018near}, fewer examples exist of real-world successes.
One example is given by~\cite{jain2019hierarchical}, in which hierarchical RL is applied to a real-world path following task. Unlike our method, this previous work uses a latent space hierarchy (as opposed to a goal-conditioned hierarchy) and trains in an end-to-end manner; thus, they are unable to take advantage of modularity with respect to domain randomization in training. 
Moreover, their real-world demonstrations focus on path following, whereas our tasks demonstrate much more complex learned behaviors.

\paragraph{Sim2Real} 
In simulated environments, model-free RL approaches to quadrupedal locomotion have long surpassed their real-world equivalents~\cite{schulman2015trust} thanks to the comparatively low cost of running large amounts of simulations~\cite{mujoco}.  
In addition to the large amounts of trial-and-error necessary for training, deploying RL in the real world also requires a robust mechanism for measuring reward, which itself can be an expensive, manual, and poorly scalable procedure~\cite{singh2019end}.
For this reason, we avoid fine-tuning in the real world~\cite{mordatch2016combining,james2019sim}, train wholly in simulation, and perform a zero-shot transfer to the real world.
To account for unknown differences between simulation and the real world we employ domain randomization, which has been used extensively to perform successful zero-shot sim2real transfer~\cite{sadeghi2016cad2rl,tobin2017domain,andrychowicz2018learning,matas2018sim}.

Traditionally, tuning the randomization parameters for a successful sim2real transfer is a manual trial-and-error process, but a few recent works attempt to improve this.
For example,~\cite{ramos2019bayessim} propose a Bayesian optimization method to optimize the distributions, which may scale poorly with the number of randomizations. Another work~\cite{mehta2019active} uses meta-learning to learn a small set of randomization settings which simultaneously maximize the utility of the learned policy and the diversity of the settings.
Our approach is markedly different from and much simpler than these existing approaches; we propose to use hierarchy to introduce modularity into the domain randomization process, which is very sensible for quadrupedal locomotion.
 One can first focus on randomizations which allow for learning a robust locomotive policy; once a basic locomotive policy is found and fixed, many of the randomizations used to train it are no longer relevant, or may be replaced with simpler variants. Learning a high-level policy on top of this basic locomotive policy will utilize distinct, and usually fewer, domain randomizations.

\section{Hierarchical Sim2Real}
\label{sec:method}
We now describe our modular approach to solving quadrupedal manipulation via locomotion.
Although this paper focuses on applications to quadrupeds, our approach is general, and can be applied to other problems involving mobile robots.  Therefore, we begin with a general overview of the method and subsequently describe the details of applying it to quadrupedal locomotion tasks.
In these descriptions we consider a standard Markov decision
process (MDP) setting~\cite{puterman1994markov}, in which each step in the environment consists of a state observation $s$, an action $a$ sampled from the policy, a reward $r$, and a next state $s'$.  We wish to find an optimal behavior policy to maximize cumulative discounted future reward, with discount given by some $0\le \gamma< 1$.

%\begin{figure}
%  \setlength{\tabcolsep}{0pt}
%  \renewcommand{\arraystretch}{0.8}
%  \begin{center}
%  	\begin{tabular}{c}
%  	\includegraphics[width=0.7\textwidth]{hrl_diagram}
%  	\end{tabular}
%  \end{center}
%  \caption{We propose to solve tasks using a hierarchical policy in which a high-level policy $\pihi$ produces high-level actions $\ahi$ which are transformed to goals $g$ that a lower-level policy $\pilo$ is trained to reach.  In quadrupedal locomotion tasks, this is a natural decomposition: the low-level policy may be trained to produce behaviors to reach various goals; i.e., $g:=(g_x,g_y)$ is a desired point and the state representation $f(s)$ is simply the $x,y$ coordinates of the quadruped. The high-level policy then solves a task by iteratively directing the low-level to a sequence of goal locations.}
%	\label{fig:hrl}
%\end{figure}

\begin{figure}
  \begin{center}
    \begin{multicols}{2}
      \begin{tabular}{ p{0.256\textwidth} p{0.744\textwidth} }
         
        \begin{minipage}{0.255\textwidth}
          \vspace{-2.2in}
          \begin{enumerate}[itemsep=1.3ex,leftmargin=0.0in]
            \scriptsize
            \item Pre-train $\pilo$ to reach randomly sampled goals $g$ from some fixed distribution; i.e, train basic locomotive policy in simulation with appropriate domain randomizations (e.g., joint frictions, random terrain).
            \item Freeze $\pilo$ and train $\pihi$ in simulation to direct $\pilo$ to solve more complex task.  Use appropriate domain randomization (e.g., noisy high-level actions).
            \item Transfer full hierarchical policy to the real world in a zero-shot fashion.
          \end{enumerate}
          \vspace{-0.15in}
        \end{minipage} &
        
        \includegraphics[width=0.7\textwidth]{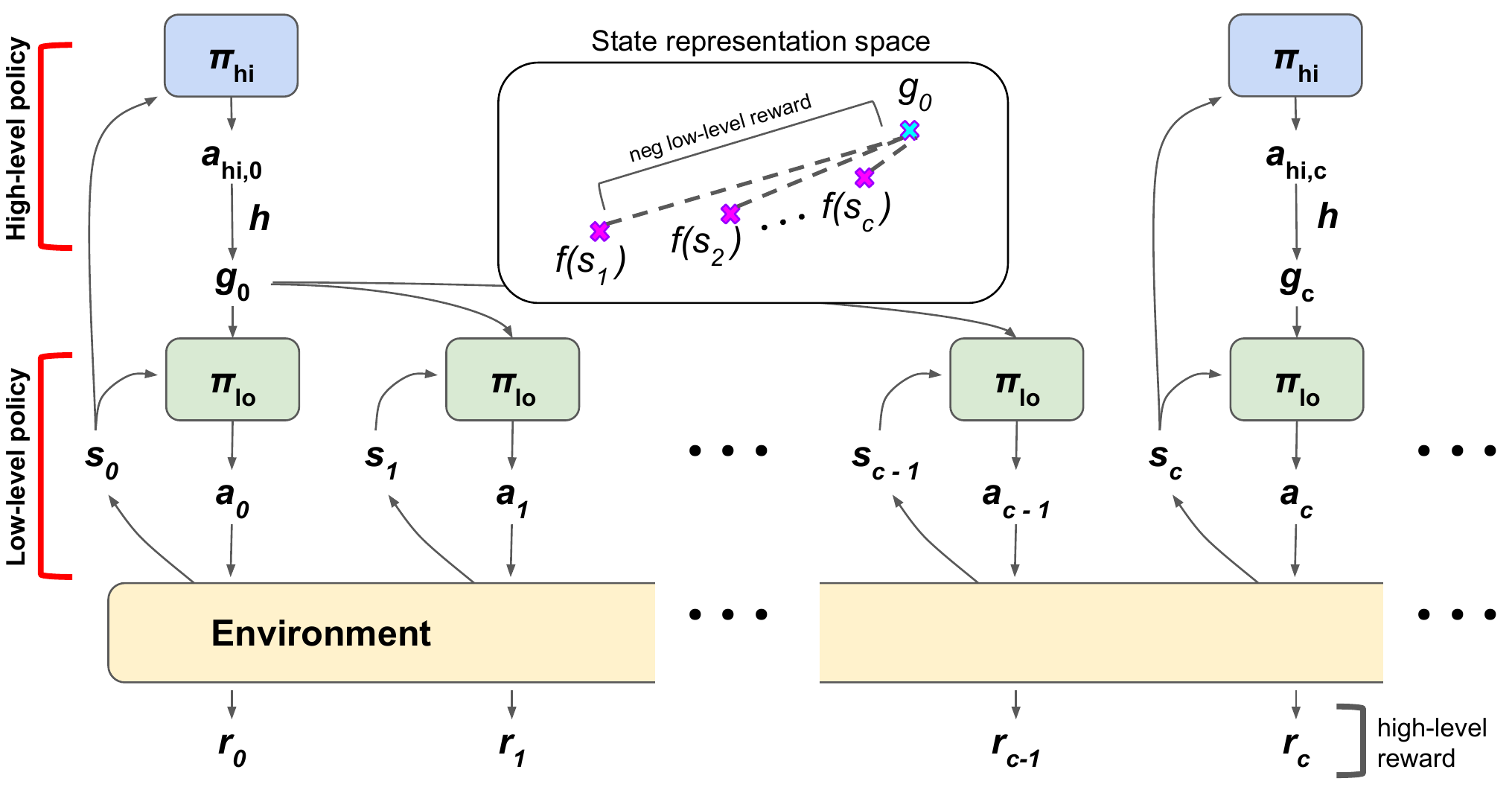} 
      \end{tabular}
    \end{multicols}
  \end{center}
  \vspace{-0.2in}
  \caption{We propose to solve tasks using a hierarchical policy in which a high-level policy $\pihi$ produces high-level actions $\ahi$ which are transformed to goals $g$ that a lower-level policy $\pilo$ is trained to reach.  In quadrupedal locomotion tasks, this is a natural decomposition: the low-level policy may be trained to produce behaviors to reach various goals; i.e., $g:=(g_x,g_y)$ is a desired point and the state representation $f(s)$ is simply the $x,y$ coordinates of the quadruped. The high-level policy then solves a task by iteratively directing the low-level to a sequence of goal locations.}
  \label{fig:hrl}
\end{figure}

\subsection{Overview}
The paradigm of hierarchical sim2real proposes to break up a task into low-level and high-level controls which can be successively learned by separate policies. For ease of exposition, we restrict our description to two-level hierarchies, although the paradigm is generalizable to hierarchies of depth greater than two. 
Taking inspiration from recent successes in simulation~\cite{hiro,nachum2018near}, we focus on {\em goal-conditioned hierarchies} in which the low-level policy is trained to reach randomly sampled {\em goal} states. The high-level policy is then trained to solve the original task by iteratively setting goals for the low-level policy.
See Figure~\ref{fig:hrl} for a visual diagram of this structure. 
In general, goals are not specified in the raw observation space but rather in a lower-dimensional representation space, with mapping from raw observations to low-dimensional representations given by some function $f$. In some settings, the function $f$ is easy to hard-code (e.g., in tasks involving quadrupedal locomotion, it is natural to choose $f(s)$ to be the $x,y$ coordinates of the robot).  In other settings, the function $f$ may be learned via unsupervised objectives~\cite{nachum2018near}.

Goal-conditioned hierarchies are desirable because low-level policies can be trained before any reward signal is observed from the original task, or even before the task is known. That is, the low-level policy is trained in an unsupervised fashion based solely on intrinsic (goal-reaching) rewards.
Accordingly, training is performed in two phases, one for each level of the hierarchy.
First, the low-level policy $\pilo(a|s,g)$ is trained to reach goals randomly sampled from some distribution $\G$. That is, we train the low-level policy via RL to maximize cumulative reward given by,
\begin{equation}
\rlo(s,a,s',g) := -D(f(s'), g)+\raux(s,a,s',g),
\end{equation}
where $D$ is some distance function, $g$ is sampled from $\G$, and $\raux$ is an optional additional auxiliary intrinsic reward.
Appropriate domain randomization should be applied during simulated training in order to encourage suitable zero-shot transfer of goal-reaching behaviors in the real world (we elaborate on the randomizations we use for quadrupedal locomotion in the next subsection).

Given a pre-trained low-level policy, the high-level policy is then trained via RL to solve the task through iterative goal-proposing.  More specifically, a high-level policy $\pihi(\ahi|s)$ is trained to take in a current observation $s$ and return a high-level action $\ahi$.  The high-level action is transformed to a goal $g:=h(\ahi)$ for the low-level via some pre-defined, fixed mapping $h$ (in simple settings, $h$ can be the identity, $g:=\ahi$).
The low-level is then run for some number of steps to generate behavior $\tau=\{(s_{i},a_{i},r_{i},s'_{i})\}_{i=0}^{c-1}$ sampled from the environment, where $s_0:=s$ and $a_{i}\sim\pilo(\cdot|s_{i},h(\ahi))$.
Following previous work~\cite{hiro}, we use a fixed $c$ ($c=10$).
The reward for the high-level is given by,
\begin{equation}
    \rhi(s,\ahi, \tau) := \sum_{i=0}^{c-1} \gamma^i r_{i}.
\end{equation}
Any standard RL algorithm may be used to optimize the policy $\pihi$ with respect to this task reward.
As for the low-level, appropriate domain randomization should be applied during high-level training.
As we will see concretely in the case of quadrupedal tasks, the domain randomization required for high-level training is often different from the domain randomization used for the low-level. It is often also much simpler than if a non-hierarchical policy is trained to solve the task from scratch.
Indeed, once one has access to a suitable goal-reaching low-level policy, this policy is fixed, and thus there is no need to employ randomizations relevant to these low-level primitives (e.g., randomizations on the joints of a walking robot or terrain surface properties).  Rather, the randomizations can focus more on high-level properties of the environment, such as agent and object locations or dimensions.

\subsection{Quadrupedal Manipulation via Locomotion}
Hierarchical sim2real as a paradigm for solving real-world tasks lends itself well to applications involving quadrupeds.
In these settings, a low-level policy may be trained to perform basic locomotion, while a high-level policy is trained to modulate these locomotive primitives in order to solve a much more complex task, involving long time-horizons, object interaction, and multiple agents.

\begin{figure}
  \setlength{\tabcolsep}{0pt}
  \renewcommand{\arraystretch}{0.8}
  \begin{center}
  \vspace{-0.1in}
  	\begin{tabular}{c}
  	\includegraphics[width=0.9\textwidth]{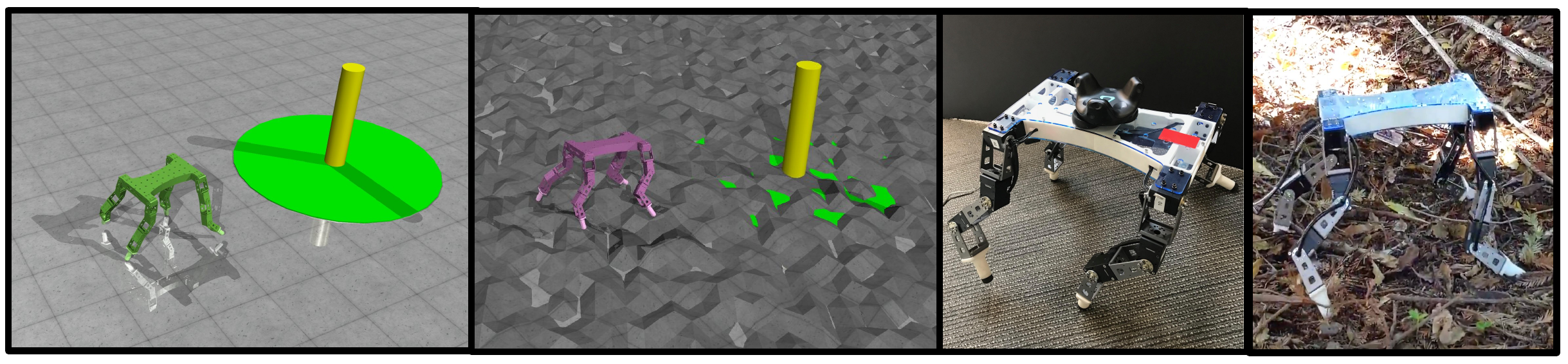}
  	\end{tabular}
  \end{center}
  \vspace{-0.1in}
  \caption{The low-level policy is trained to perform simple goal-reaching in simulation (left). We apply a variety of domain randomizations, including randomized height fields, which we found to be most helpful (middle).  These randomizations lead to a robust locomotion policy in a variety of real-world environments, including both indoor and outdoor terrain (right).}
  \vspace{-0.2in}
	\label{fig:low-level}
\end{figure}

\subsubsection{Training}
\label{sec:training}
We train the low-level to reach goals in $x,y$ space; i.e., $f(s)$ returns the $x,y$ (ground) coordinates of the torso of the quadruped. 
During low-level training we sample goals $(g_x,g_y)$ as,
\begin{equation}
    (g_x,g_y) := (u\cos v, u\sin v), ~\text{where}~ u\sim \uniform(1, 2), v\sim\uniform(-0.5, 0.5).
\end{equation}
We reward the low-level with negative L2 distance to the goal; i.e., $D:=||\cdot||_2$. To encourage better walking behavior, we define $\raux$ to augment the reward with an additional three terms,
\begin{equation}
    \label{eq:raux}
    \raux(s,a,s',g) := r_{\mathrm{upright}}(s') + r_{\mathrm{heading}}(s',g) + r_{\mathrm{bonus}}(s', g).
\end{equation}
The first term $r_{\mathrm{upright}}$ encourages the quadruped to remain upright; i.e., penalizes it when the torso's $z$-coordinate is too low.
The second term $r_{\mathrm{heading}}$ encourages the quadruped to face the goal point; i.e., rewards it for having a high normalized dot-product between $g-s'$ and the torso's heading.  Lastly, the third term $r_{\mathrm{bonus}}$ augments the reward with a bonus when the quadruped is within $0.5$m of the goal.
We run episodes for length 40, corresponding to 4 seconds of real-world time, terminating the episode if the quadruped falls. 

Once the low-level is trained, we fix it and train the high-level to direct the low-level to solve a more complex task over a longer horizon (200 steps in simulation, 20 seconds in the real world).
In the case of a single mobile agent, the high-level actions are 2-dimensional.  A high-level action $\ahi:=(a_{\mathrm{hi},0},a_{\mathrm{hi},1})$ is transformed to a low-level goal via the polar mapping,
\begin{equation}
    \label{eq:polar}
    h(\ahi) := (u(\ahi) \cos v(\ahi), u(\ahi)\sin v(\ahi)),
\end{equation}
where, $u,v$ are linear functions: $u(\ahi) := 0.5 + 0.3\cdot a_{\mathrm{hi},0}$; $v(\ahi) := 0.2\cdot a_{\mathrm{hi},1}$.

To ensure that the low-level generalizes to potentially very different environments, when activating the low-level policy, we zero-out the components of each observation $s$ which correspond to the 6 positional and rotational degrees of freedom of the torso.  Thus, high-level actions always correspond to {\em relative} goals (similar to previous work in simulation~\cite{hiro}).

In multi-agent environments we allow the high-level policy to control multiple mobile agents concurrently in order to coordinate their behavior.  In these settings, the high-level actions are $2n$-dimensional, where $n$ is the number of quadrupedal agents, and $h$ is defined analogously to Equation~\ref{eq:polar}. The modularity enabled through hierarchy in these multi-agent settings is clear. Rather than interact with the high-dimensional action space of a multi-agent environment directly, the imposition of hierarchy allows us to first train a single low-level policy and then train a high-level which effectively treats each agent as a point mass.  Hierarchy here also makes the domain randomization problem more tractable, avoiding the duplication that will otherwise be necessary for training two agents from scratch. 

We learn each policy by fitting a neural network using natural policy gradient~\cite{mjrl1, mjrl2}. 
Each policy is a feed-forward neural network with two hidden layers, each of dimension 32 and using $\tanh$ activations.
We train the low-level for 15000 iterations using batches of 100 trajectories. The high-level is then trained for 5000 iterations using batches of 200 trajectories.
%The resulting policy is able to successfully transfer to the real world in a zero-shot fashion.

\begin{figure}
  \setlength{\tabcolsep}{0pt}
  \renewcommand{\arraystretch}{0.8}
  \begin{center}
  \vspace{-0.1in}
  	\begin{tabular}{c}
  	\includegraphics[width=0.8\textwidth]{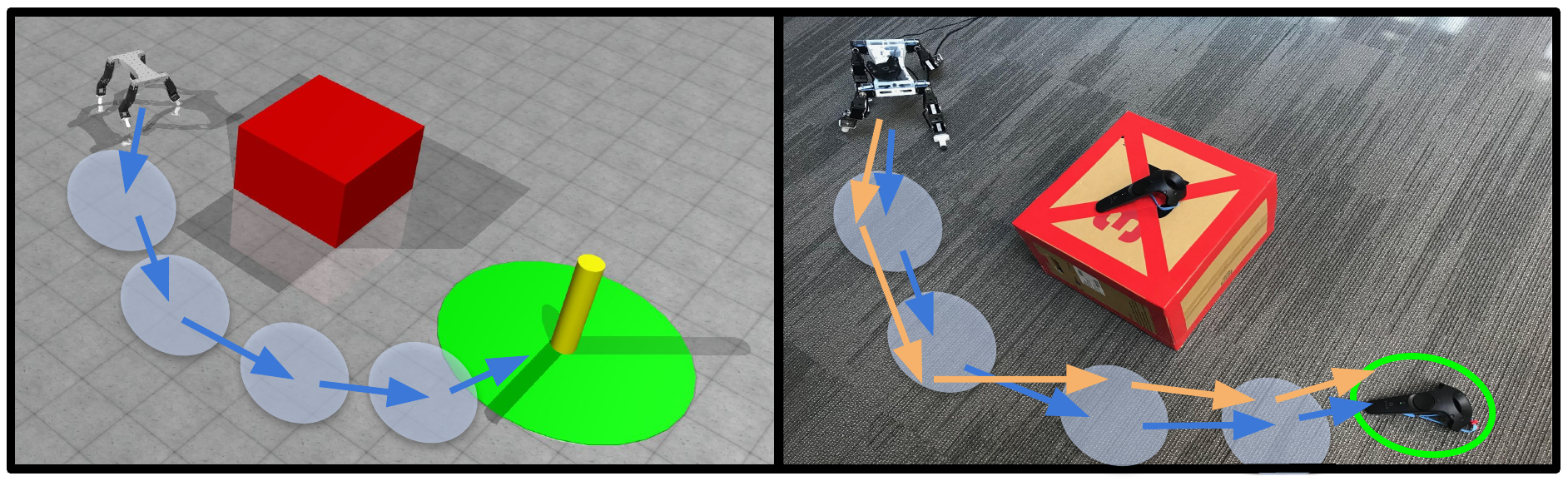}
  	\end{tabular}
  \end{center}
  \vspace{-0.1in}
  \caption{The high-level policy directs the agent by setting relative position goals.  In this way, it combines locomotive primitives to solve a more complex task, as in this example trajectory for the \textit{Avoid} task. To account for potential unknown gaps between low-level behavior in simulation and reality (blue vs. orange arrows), during training we pollute the high-level actions with random noise.}
  \vspace{-0.2in}
	\label{fig:high-level}
\end{figure}

\subsubsection{Domain Randomization}
\label{sec:domain}
In addition to simplifying the training of complex behaviors, hierarchy also introduces modularity into the domain randomization process.
For low-level domain randomization, we randomize a number of parameters commonly used in sim2real methods~\cite{andrychowicz2018learning}: joint damping, joint friction, actuator gain, total mass (gravitational force), and surface friction.  As in previous works, we also apply random forces and torques to the torso (randomly applying anew every 10 simulated steps) and use sticky actions (repeat the previous action) with probability $0.2$.
Finally, we replace the floor with a randomly generated height field; i.e., we replace the terrain with a randomly generated triangular mesh (see Figure~\ref{fig:low-level}).
In MuJoCo~\cite{mujoco}, this corresponds to simply adding an \texttt{hfield} asset to the model and randomizing its parameters.
We are not aware of previous work which utilizes this randomization.
In our experience, we found this randomization to be the {\em most crucial} for producing robust walking gaits, and hope that future quadrupedal locomotion research can make further use of this finding.
See the appendix for more randomization details.

Once the low-level is sufficiently trained, it is effectively a robust moving point mass from the perspective of the high-level, so the next stage of training requires significantly fewer domain randomizations.
Still, although the pre-trained low-level policy is robust, it is not perfect, and there remains a non-negligible gap between its behavior in simulation and the real world.  For example, a `walk forward' command may result in real-world behavior that is slightly biased to the left or right due to hardware imperfections or terrain variability; see Figure~\ref{fig:high-level}.
Rather than re-use the various randomizations employed for low-level training, we instead use a single, simple randomization to account for this gap: We add noise to the high-level actions (uniformly sampled between -1 and 1) before passing them to the low-level.  This is another example where the benefits of hierarchy for sim2real applications are concrete. Training a non-hierarchical policy on these more complex, longer horizon tasks would potentially involve the cumbersome tuning of various short-term and long-term randomization parameters. In contrast, when hierarchy is imposed, the quadruped is simply a noisy point mass from the perspective of the high-level policy, and thus appropriate domain randomization is much simpler to apply.
In the multi-agent task, in addition to adding noise to the high-level actions, we also randomize the object dimensions.

With these domain randomizations, the resulting hierarchical policy is able to successfully transfer to the real world in a zero-shot fashion.

\vspace{-0.1in}
\section{Experiments}
\label{sec:exp}
\vspace{-0.1in}
We present experimental results of applying hierarchical sim2real on several real-world tasks, building up to a demonstration of coordinated multi-agent object manipulation.
We begin by describing the hardware and the tasks on which we evaluate.

\vspace{-0.1in}
\subsection{Hardware}
\label{sec:hardware}
\vspace{-0.1in}
For our quadrupedal robot we use D'Kitty, a robust and affordable locomotion platform (4200 USD) designed and built in-house.  The robot consists of a 3D-Printed torso and four legs, each comprised of three Dynamixel Servo Motors that provide feedback on joint angle and angular velocity.
We attach an HTC Vive tracker to the torso to further observe the global position and orientation of the robot.  In tasks involving objects (blocks), we additionally use an HTC Vive controller attached to the object to observe its global position and orientation.

\vspace{-0.1in}
\subsection{Tasks}
\vspace{-0.1in}
We consider three tasks (see Figure~\ref{fig:tasks} for a visual diagram):
\begin{itemize}[leftmargin=*,noitemsep,topsep=0pt,parsep=0pt,partopsep=0pt]
    \setlength\itemsep{0pt}
    \item {\bf Avoid}. The quadruped must walk to a target location while avoiding a block object.  We provide the agent a reward at each step consisting of (1) a term measuring negative L2 distance of the agent to the target, (2) a bonus for being within $0.5$m of the target location, (3) a term which penalizes the agent for being within $0.5$m of the block location, and (4) a term which encourages the agent to be upright (same as $r_{\mathrm{upright}}$ in Equation~\ref{eq:raux}).
    \item {\bf Push}.  The quadruped must push a block object to a desired target location.  We provide the agent a reward at each step consisting of (1) a term measuring negative L2 distance of the agent to the block, (2) a term measuring negative L2 distance of the block to the target, (3) a bonus for when the block is within $0.5$m of the target location, and (4) a term which encourages the agent to be upright (same as $r_{\mathrm{upright}}$ in Equation~\ref{eq:raux}).
    \item {\bf Coordinate}.  Two quadrupeds must coordinate to push a long block such that the two ends of the block match two desired target locations.  We provide the agent a reward at each step consisting of (1) the sum of the negative L2 distances of each end of the block to the desired target locations, (2) a bonus for when both ends of the block are withing $0.3$m of the desired target locations, and (3) a term which encourages both agents to be upright (the sum of $r_{\mathrm{upright}}$ for each agent).
\end{itemize}
Each task is run for 200 steps (corresponding to 20 seconds in the real world) without termination.  The task observations include the joint angles, joint angular velocities, global position, and global orientation of each quadruped.  In \textit{Avoid} and \textit{Push}, the observations include positions of the target and block locations relative to the quadruped. In \textit{Coordinate}, the observations include the orientation of the block, the positions of each target location relative to each end of the block, and the positions of each end of the block relative to each quadruped.

In simulation, the positions of the quadrupeds, objects, and target locations are initialized randomly.  In the real world, we measure the success rate of the hierarchical policy on the three tasks (whether it reached the target or not) based on fixed initializations.  See the appendix for further details.

\begin{table}[t]
    \begin{center}
    \begin{tabular}{ | c | c | c | c | c |}
    %\begin{tabular}{c c }
        \hline
        %\multicolumn{2}{c}{\bf Average Reward}\\
        & \begin{tabular}{@{}c@{}}Hierarchical \\ Sim2Real\end{tabular} 
        & \begin{tabular}{@{}c@{}}No high-level \\ randomization\end{tabular} 
        &
        \begin{tabular}{@{}c@{}}No hierarchy, \\ with randomization\end{tabular} 
        &
        \begin{tabular}{@{}c@{}}No hierarchy, \\ no randomization\end{tabular} \\
        \hline
        Avoid & {\bf 73.3 $\pm$ 6.7\%} & $56.7\pm18.6\%$ & $0\pm0\%$ & $0\pm0\%$ \\
        Push & {\bf 73.3 $\pm$ 12.0\%} & $70.0\pm5.8\%$ & $0\pm0\%$ & $20.0\pm5.8\%$ \\
        %\hline
        Coordinate & {\bf 90.0 $\pm$ 5.7\%} & $80.0\pm15.3\%$ & $0\pm0\%$ & $6.7\pm3.3\%$ \\
        \hline
    \end{tabular}
    \end{center}
\caption{
We present success rates (with standard error) on the three tasks comparing our method (hierarchical sim2real) to a number of ablations. Each percentage is the result of 30 attempts -- ten attempts per three independently trained models.  These ablations show that hierarchy is crucial to success on these tasks.  With no hierarchy (right two ablations), success rates fall precipitously. Once hierarchy is employed, domain randomization on the high-level (uniform noise added to high-level actions) can provide a modest improvement.
}
\vspace{-0.1in}
\label{tab:high-level}
\end{table}

\begin{table}[t]
    \begin{center}
    \begin{tabular}{ | c | c | c | c |}
    %\begin{tabular}{c c }
        \hline
        %\multicolumn{2}{c}{\bf Average Reward}\\
        & \begin{tabular}{@{}c@{}}All randomizations\end{tabular} 
        & \begin{tabular}{@{}c@{}}No height field\end{tabular} 
        &
        \begin{tabular}{@{}c@{}}No randomizations\end{tabular} 
        \\
        \hline
        Distance Travelled & {\bf 1.26 $\pm$ 0.43} & $0.85\pm0.36$ & $0.44\pm0.14$ \\
        Percentage of Falls & {\bf 0\%} & {\bf 0\%} & $80\%$ \\
        \hline
    \end{tabular}
    \end{center}
\caption{We present the performance of the low-level policy at basic locomotion.  We run the low-level policy to walk forwards for five seconds, and show the average distance travelled (in meters, with standard error) and the rate of falling.  Each statistic is the result of 30 attempts -- ten attempts per three independently trained models. Without randomizations, the policy is unusable in the real world, due to high rate of falling.  With randomizations, the rate of falling goes to zero.  We find the most impactful randomization to be the height field, which improves the average distance travelled by about 50\%.
}
\vspace{-0.1in}
\label{tab:low-level}
\end{table}

\vspace{-0.1in}
\subsection{Results}
\label{sec:results}
\vspace{-0.1in}
After training the low-level and high-level in simulation, we evaluate the full hierarchical policy in the real world.  We present the results in Table~\ref{tab:high-level}.  
To ensure confidence in our metrics, we measure success over 30 trials, where each set of 10 is a separately trained model.

Our results present the hierarchical sim2real approach compared to a number of ablations.  
In `No hierarchy, with randomization' and `No hierarchy, no randomization,' we train non-hierarchical policies in simulation with and without randomization, respectively.  While both of these policies perform poorly in the real world, the variant with domain randomization surprisingly performs even worse. Although not shown (see the appendix), we note that {\em in simulation} the `No hierarchy, no randomization' ablation can reasonably solve all the tasks, while the `No hierarchy, with randomization' cannot.  We therefore conclude that the randomizations which allow for learning robust basic locomotion are too aggressive to learn more complex locomotion, and lead to an overly conservative policy, as shown by our results. This is strong evidence of the need to apply hierarchy in a sim2real context.  Hierarchy introduces modularity into the sim2real process with real-world benefits.

The two hierarchical ablations -- `Hierarchical Sim2Real' (our proposed method) and `No high-level randomization' -- achieve success rates much higher than either of the non-hierarchical ablations. Once a low-level basic locomotive policy is learned, the domain randomizations necessary for low-level training can be discarded, and a high-level policy can be easily trained.  Even with no randomizations (`No high-level randomization'), high-level training in simulation achieves good zero-shot results in the real world.  When adding basic high-level randomizations (as described in Section~\ref{sec:domain}), we are able to further improve the success rates, reaching 90\% success on \textit{Coordinate}.

We provide further insight into low-level training in Table~\ref{tab:low-level}. The randomizations described in Section~\ref{sec:domain} lead to a robust locomotive policy.
In comparison, a policy trained with no randomizations may perform well in simulation, but in the real world often falls shortly after initiating movement.  We found height field randomization to be the most useful element leading to a robust policy.  It alone provides a $50$\% boost in performance.
\section{Conclusion}
\label{sec:conc}
We have presented successful zero-shot transfer of policies trained in simulation to perform difficult locomotion and manipulation via locomotion tasks.
The key to our method is the imposition of hierarchy, which introduces modularity into the domain randomization process and enables the learning of increasingly complex behaviors.  We intend to continue our efforts in future work, further exploiting hierarchical structure to demonstrate more difficult tasks with a greater number of cooperating agents.
%===============================================================================

% The maximum paper length is 8 pages excluding references and acknowledgements, and 10 pages including references and acknowledgements

\clearpage
% The acknowledgments are automatically included only in the final version of the paper.
\acknowledgments{
We thank Chad Richards, Byron David, Matt Neiss, Krista Reymann, Ben Eysenbach, Sergey Levine, and the rest of Robotics at Google for helpful thoughts and discussions.
}

%===============================================================================

\bibliography{ref}  % .bib

\newpage
\appendix
\section{Task Details}
For \textit{Avoid} we use a simulated block of dimensions $0.6\times0.6\times0.4$ meters.  The target location is set at the beginning of each episode as $(r_T\cos\theta_T, r_T\sin\theta_T)$, where $r_T\sim\uniform(1.5, 2.5)$ and $\theta_T\sim\uniform(-1, 1)$.  The block location is then set to $(r_B\cos\theta_B,r_B\sin\theta_B)$ where $r_B=\max\{0.6, r_T\cdot\uniform(0.3, 0.8)\}$ and $\theta_B=\theta_T + \uniform(-0.5, 0.5)$.

For \textit{Push} we use a simulated block of dimensions $0.6\times0.6\times0.4$ meters.  The target location is set at the beginning of each episode as $(r_T\cos\theta_T, r_T\sin\theta_T)$, where $r_T\sim\uniform(1.5, 2.5)$ and $\theta_T\sim\uniform(-1, 1)$.  The block location is then set to $(r_T\cos\theta_T + r_B\cos\theta_B,r_T\sin\theta_T + r_B\sin\theta_B)$ where $r_B\sim\uniform(0.6, 1.2)$ and $\theta_B\sim\uniform(\pi/3,5\pi/3)$.  If this results in the block and agent being within $0.5$m, we re-sample.

For \textit{Coordinate} we use a simulated block of dimensions $0.3\times1.4\times0.3$ meters.  The block location is initialized to always be in front of the agents.  The two target locations are then set to be $(r_T\cos(\theta_T) \pm 0.55 \cos \theta_B, r_T\sin \theta_T \pm 0.55 \sin \theta_B)$, where $r_T\sim\uniform(1, 1.5)$, $\theta_T\sim\uniform(-1, 1)$, $\theta_B\sim \theta_T + \pi/2 + \uniform(-0.5, 0.5)$.

During real-world evaluations, we initialize the tasks as follows:
\begin{itemize}[leftmargin=*,noitemsep,topsep=0pt,parsep=0pt,partopsep=0pt]
    \setlength\itemsep{0pt}
    \item {\bf Avoid}. The quadruped is initialized facing a box 1m away.  Taking the robot's heading as the x-axis, we place the target location at $(2,\pm0.5)$ (i.e., have the time at $(2,0.5)$ and half the time at $(2,-0.5)$.  Success is measured as reaching within 0.5m of the target without moving the box.
    \item {\bf Push}.  The quadruped, box, and target initializations are the same as for Avoid.  Success is measured as pushing the box within 0.5m of the target location.
    \item {\bf Coordinate}.  The two quadrupeds are initialized facing a long, horizontal box.  The target location is set to be a 1.5m frontwards translation of the box.  Success is measured as both ends of the box reaching within 0.3m of the target position of the ends of box.
\end{itemize}

\section{Domain Randomization}
The code snippets below show the specific randomizations we employed for low-level training.

\lstset{language=python,frame=single}
\begin{lstlisting}[frame=tb,language=Python,basicstyle=\footnotesize\ttfamily,caption=Dynamics model randomizations for low-level training.]
dof0 = 6  # six global observations
dofn = dof0 + 12  # twelve joints
self.sim.model.dof_damping[dof0:dofn] = \ 
    self.np_random.uniform(low=0.9, high=1.1)
self.sim.model.dof_frictionloss[dof0:dofn] = \ 
    self.np_random.uniform(low=0.001, high=0.005)
functions.mj_setTotalmass(
    self.sim.model, 
    self.np_random.uniform(low=1.6, high=2.0))
self.sim.model.geom_friction[:] = self.np_random.uniform(
    low=[0.8, 0.003, 0.00005], high=[1.2, 0.007, 0.00015])
self.sim.model.actuator_gainprm[:, 0] = self.np_random.uniform(
    low=4, high=6)
self.sim.model.actuator_biasprm[:, 1] = \ 
    -self.sim.model.actuator_gainprm[:, 0]

# regenerate height map and upload to GPU
self.sim.model.hfield_data[:] = self.np_random.uniform(
    low=0, high=.050, size=np.shape(self.sim.model.hfield_data))
if (len(self.sim.render_contexts) > 0):
    functions.mjr_uploadHField(self.sim.model,
                               self.sim.render_contexts[0].con, 0)
\end{lstlisting}

%\begin{lstlisting}[frame=tb,language=Python,basicstyle=\footnotesize\ttfamily,caption=Force and torque randomizations for low-level training.]
%if self.time_step % 10 == 1:
%    if self.last_force is not None:
%        functions.mj_applyFT(self.sim.model, self.sim.data,
%                             -self.last_force, -self.last_torque,
%                             np.zeros([3]),  # point relative to body
%                             self.torso_sid,  # body
%                             self.sim.data.qfrc_applied)
%    fscale = 1
%    tscale = 0.1
%    force = self.np_random.uniform(low=-fscale, high=fscale, size=3)
%    torque = self.np_random.uniform(low=-tscale, high=tscale, size=3)
%    functions.mj_applyFT(self.sim.model, self.sim.data,
%                         force, torque,
%                         np.zeros([3]),  # point relative to body
%                         self.torso_sid,  # body
%                         self.sim.data.qfrc_applied)
%    self.last_force = force
%    self.last_torque = torque
%\end{lstlisting}

\section{Training and Ablation Details}
By default, each policy is a feed-forward neural network with two hidden layers, each of dimension 32 and using $\tanh$ activations.
For the non-hierarchical ablations we increase the hidden dimension to 64.  
For the non-hierarchical ablation with randomization, we apply all the randomizations used by default for low-level training and additionally a small (uniform between -0.1 and 0.1) noise on the torso $x,y$ coordinate observations.

\section{Simulation Results}

\begin{table}[h]
    \begin{center}
    \begin{tabular}{ | c | c | c | c | c |}
    %\begin{tabular}{c c }
        \hline
        %\multicolumn{2}{c}{\bf Average Reward}\\
        & \begin{tabular}{@{}c@{}}Hierarchical \\ Sim2Real\end{tabular} 
        & \begin{tabular}{@{}c@{}}No high-level \\ randomization\end{tabular} 
        &
        \begin{tabular}{@{}c@{}}No hierarchy, \\ with randomization\end{tabular} 
        &
        \begin{tabular}{@{}c@{}}No hierarchy, \\ no randomization\end{tabular} \\
        \hline
        Avoid & {\bf 86.5 $\pm$ 5.8\%} & $76.7\pm3.3\%$ & $0\pm0\%$ & $33.3\pm20\%$ \\
        Push & {\bf 93.3 $\pm$ 8.8\%} & $90.0\pm11.5\%$ & $0\pm0\%$ & $86.7\pm6.7\%$ \\
        %\hline
        Coordinate & $80.0\pm3.3$\% & $66.7\pm3.3\%$ & $0\pm0\%$ & {\bf 86.7 $\pm$ 30\%} \\
        \hline
    \end{tabular}
    \end{center}
\caption{
We present success rates (with standard error) on the three tasks in simulation. Each percentage is the result of 30 attempts -- ten attempts per three independently trained models.  In simulation we find that hierarchy helps, but less so than in the real world, especially on the \textit{Coordinate} task, for which a non-hierarchical policy performs best.  This is potentially because our low-level policies are not trained in the presence of blocks.  We also see that learning a policy from scratch (no hierarchy) with domain randomization is futile. In order to successfully transfer learned behaviors to the real world, both hierarchy and domain randomization are necessary.
}
\vspace{-0.1in}
\label{tab:sim-results}
\end{table}

\end{document}